\documentclass{article}
\usepackage{spconf}
\usepackage{amssymb}
\usepackage{array}
\usepackage{multirow}
\usepackage{graphicx}

\title{Dataset Balancing Can Hurt Model Performance}
\name{
    R. Channing Moore,
    Daniel P. W. Ellis,
    Eduardo Fonseca,
    Shawn Hershey,
    Aren Jansen,
    Manoj Plakal}
\address{Google Research, 
         Mountain View, CA, and New York, NY, USA\\
         \textit{\{channingmoore, dpwe, efonseca, shershey, arenjansen, plakal\}@google.com}}

\begin{document}
\ninept
\maketitle
\begin{abstract}
Machine learning from training data with a skewed distribution of examples per class can lead to models that favor performance on common classes at the expense of performance on rare ones. AudioSet has a very wide range of priors over its 527 sound event classes. Classification performance on AudioSet is usually evaluated by a simple average over per-class metrics, meaning that performance on rare classes is equal in importance to the performance on common ones. Several recent papers have used dataset balancing techniques to improve performance on AudioSet. We find, however, that while balancing improves performance on the public AudioSet evaluation data it simultaneously hurts performance on an unpublished evaluation set collected under the same conditions.  
By varying the degree of balancing, we show that its benefits are fragile and depend on the evaluation set. We also do not find evidence indicating that balancing improves rare class performance relative to common classes. We therefore caution against blind application of balancing, as well as against paying too much attention to small improvements on a public evaluation set.

\end{abstract}

\section{Introduction}
Recent advances in transformer models have improved the state of the art in natural-language processing and in image and audio classification machine learning (ML) tasks \cite{vaswani2017attention, dosovitskiy2020image, gong21b_interspeech}. These models have large capacity and can make use of large quantities of labeled training data. The largest set of publicly-available labels for audio event classification, AudioSet \cite{gemmeke2017audio}, is small compared to some of the datasets commonly used to train image models like ImageNet-22k \cite{russakovsky2015imagenet} or JFT \cite{hinton2015distilling}. This makes pretraining and dataset manipulations critical.

The AudioSet task involves recognizing 527 sound event classes within $\sim$10-second audio clips. These classes display large variations in their intrinsic difficulty as well as in their prevalence in the 2M clips of the training set: The most common, \textit{music}, occurs roughly 15,000 times more frequently than the rarest (\textit{toothbrush}).  

AudioSet performance is usually reported as an unweighted average of per-class metrics (such as average precision) across all classes. To improve the performance on the rarer classes, several authors have implemented \textit{balanced sampling} in training, where samples of the less-common classes are repeated to improve their representation during training. This has shown improvements in overall performance when combined with data augmentation techniques \cite{jeong2018audio, kong2020panns, gong2021psla, gong21b_interspeech}.

In this work, we compare the impact of this approach on the public AudioSet evaluation set to a separate, internal evaluation set collected at the same time.  This second set has a different, more skewed class distribution. Unexpectedly, we find that balancing the dataset actually worsens performance on this second evaluation set.

We present this as evidence that class balancing, expected to improve allocation of model capacity and learned prior, may not be significantly useful in the AudioSet domain---and that previously-reported benefits may reflect immaterial peculiarities of the widely-used evaluation set. 

\begin{figure}[t]
\centering
\includegraphics[scale=1.0]{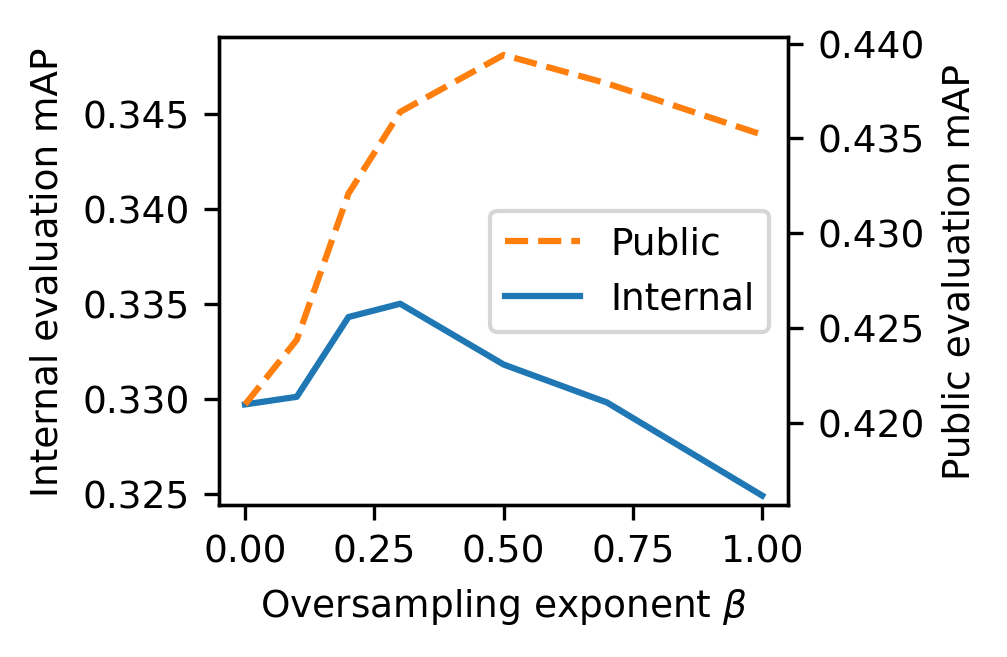}
\caption{mAP as a function of oversampling exponent $\beta$ for Public and Internal evaluation sets. The two traces are plotted with the same vertical scaling, but are vertically offset to align performance without balanced sampling ($\beta=0$).}
\label{fig:MAP}
\end{figure}

\subsection{Class imbalance in AudioSet partitions}
The creators of AudioSet\footnote{Authors Moore, Hershey, Plakal, Jansen, and Ellis were among the creators of AudioSet.} produced balanced training and evaluation partitions by picking 60 examples for the rarest class, then adding examples of the next-rarest class until there were 60 examples of that class, and so forth. These balanced sets still have significant imbalance because the examples have multiple labels and common classes co-occur with uncommon ones. For example, of the 60 \textit{toothbrush} training examples, 41 also have the label \textit{speech}, 10 have the label \textit{music}, and 8 have both. In this way, common classes remain over-represented even after these attempts to balance the dataset.

\subsection{Class imbalance and performance}
Class imbalance could in principle skew training of machine-learning models \cite{he2009learning} (although see the discussion of Figure 7 in \cite{gong2021psla} for other factors). Although class prior is not directly correlated with per-class performance, reducing imbalance could help. In addition to the improved performance on audio classification mentioned above, similar techniques are also reported under the name \textit{class-aware sampling}, primarily in the image domain. \cite{shen2016relay, gao2018solution}. We are not aware of any work which has used balanced sampling for automatic speech recognition.

\section{Methods}

\subsection{Quantifying class imbalance}
With $N$ total examples and $K=527$ total classes in AudioSet, we define $j \in \{1..N\}$ as the sample index and $k \in \{1..K\}$ as the class index. Given the label matrix $C \in \{0,1\}^{N{\times}K}$ with entries $c_{jk}$ we define the prior $p_k$ for each class $k$ as

\begin{equation}
    p_k = \frac{1}{N} \sum_{j=1}^N{c_{jk}} \equiv \frac{N_k}{N},
\end{equation}

\noindent where $N_k = \sum_{j=1}^N c_{jk}$ is the number of examples for which class $k$ is marked as present.

We can measure class imbalance using the \textit{imbalance ratio} \cite{kim2021imbalanced}, which we will denote as $\rho$,

\begin{equation}
    \rho = \frac{\max_k(p_k)}{\min_k(p_k)} \equiv \frac{\max_k(N_k)}{\min_k(N_k)}.
\end{equation}

This measure gives a simple, intuitive sense of how unbalanced the dataset is, but it only considers two samples from the prior distribution, the head and tail classes. To better measure the degree of imbalance across the entire dataset we compute the Gini coefficient \cite{gini1912variabilita}, a common measure of how far data differ from a uniform distribution. There are more than one label per example---roughly 2 on average---so we compute this measure based on the fraction of total labels rather than directly from the prior distribution.

If $p_k$ is list of the per-class priors sorted in ascending order, the Gini coefficient is equal to

\begin{equation}
    1 - \frac{2 \cdot \sum_{k=1}^K \sum_{l=1}^{k} p_l}{K \cdot \sum_{k=1}^{K}p_k}.
\end{equation}

\subsection{AudioSet}

\subsubsection{Current version of the dataset}
The version of AudioSet we used for this paper consists of the 1,728,000 training clips and 16,591 evaluation clips that remain available. The statistics of this version of the dataset are relatively unchanged from the original publication: the most common class is still \textit{music} (prior 0.482), and the least common is \textit{toothbrush} (prior $2.66\times10^{-5}$).

\subsubsection{Validation dataset}
We reserved examples from the AudioSet training dataset as a validation set. We chose these examples by a similar procedure to the evaluation set: we required a minimum of 5 examples of each class. The current version of this set has 1715 clips.

\subsubsection{Internal evaluation dataset}
We have labels for a further set of 22,573 clips that were not published. We also evaluate on the union of this set with the public evaluation set (a total of 39,164 clips) and refer to this combined set as the \textit{internal evaluation} set. 
We found and annotated the labels for the additional internal evaluation clips at the same time as the public AudioSet clips, using the same procedures. %

\subsubsection{Class prior imbalance}
Table \ref{tab:priors} summarizes the class imbalance in the various data partitions. The 
training set we used has substantially the same class imbalance as the published training set when measured by the Gini coefficient. The two evaluation sets are more balanced, and the validation set falls in between the two evaluation sets.

\begin{table}[t]
    \centering
    \begin{tabular}{c|c|c}
        {}  & Imbalance ratio    & Gini coefficient \\
        \hline
        Published train  & 15,009  & 0.83 \\
        Current train   & 18,102    & 0.83 \\
        Validation  & 275   & 0.45 \\
        Public evaluation & 181   & 0.39 \\
        Internal evaluation & 367   & 0.61
    \end{tabular}
    \caption{
        Prior imbalance statistics for data partitions. We computed statistics for \textit{published train} from the original list of labels published with \cite{gemmeke2017audio}. All other statistics are computed from the dataset version used in this paper.}
    \label{tab:priors}
\end{table}

\subsection{Model architecture and training}
We replicated the AST model architecture of \cite{gong21b_interspeech} in TensorFlow and pretrained it on a video-level tag prediction task with a fixed ontology of 10k classes (not necessarily audio-related) on a collection of 50 million 10-second audio clips from internet videos. We then refined the full model on AudioSet. We performed refinement using binary cross-entropy loss, Adam optimizer \cite{kingma2014adam}, and a batch size of 1024. We did not use the learning-rate warmup, learning-rate decay, weight decay, weight averaging, or model ensembling described in \cite{gong21b_interspeech}.

\subsection{Balancing}
We balanced the training dataset by repeating examples of low-prior classes, repeating examples $M_j \in \mathbb{Z^+}$ times on each pass through the training data. Defining $m_j \in \mathbb{R^+}$ as

\begin{equation}
    m_j = \max_{k:c_{jk}=1}\left(\frac{1}{p_k}\right)^\beta,
\end{equation}

\noindent we compute the oversampling factor $M_j$ in terms of $m_j$ as

\begin{equation}
    M_j = \textrm{round}\left(\frac{m_j}{\min_j(m_j)}\right).
\end{equation}

Our balancing scheme modifies previous work by adding the parameter $\beta$, which we term the \emph{oversampling exponent}. This allows us to choose oversampling schemes that lie between the raw, unbalanced AudioSet dataset at $\beta=0$ and the full oversampling used by \cite{gong21b_interspeech}, \cite{gong2021psla}, and \cite{kong2020panns} at $\beta=1$. At $\beta=0$, $M_j \equiv 1$ for all examples. At $\beta=1$, $M_j$ ranges from 1 for examples marked with only the two most common classes \emph{speech} and \emph{music} to 18,102 for examples marked with \textit{toothbrush}. Table \ref{tab:oversampled_priors} shows the dataset statistics from this partial balancing.

We did not implement the per-batch balancing scheme of \cite{kong2020panns}. Our training batch size of 1024 is large enough that the oversampling procedure ensures that most classes are represented in any given batch. The rarest class in the fully-balanced dataset has a frequency of 1 in 435 examples, showing up on average more than twice in each batch.

\begin{table}[t]
    \centering
    \begin{tabular}{c|c|c}
        Oversampling exponent $\beta$  & Imbalance ratio    & Gini coefficient \\
        \hline
        0.0     & 18102     & 0.83 \\
        0.1		& 8282		& 0.79 \\
        0.2		& 4916		& 0.75 \\
        0.3		& 2746		& 0.71 \\
        0.5		& 1032		& 0.63 \\
        0.7		& 443		& 0.56 \\
        1.0		& 143		& 0.47 \\
    \end{tabular}
    \caption{
        Prior imbalance statistics for oversampled AudioSet training set.}
    \label{tab:oversampled_priors}
\end{table}

\subsection{Augmentation}

We used SpecAugment \cite{park2019specaugment} with the same parameters as \cite{gong21b_interspeech}, and applied MixUp \cite{zhang2017mixup} in the energy domain with $\alpha$=10 for all of our experiments. We combined examples within batches rather than drawing a separate batch of augmentation examples. 
Our work differs from \cite{gong21b_interspeech} in that we only used a MixUp rate of 1, meaning that all training examples are subject to MixUp. 

\section{Evaluation}

\subsection{Metrics}
The AST model produces one score per clip for each of the 527 classifier outputs. We computed mean average precision (mAP) and the area under the ROC curve (AUC) for each classifier and averaged them using equal weight for each class (macro-averaging).
We then computed $d^\prime$ from the macro average AUC as described in \cite{hershey2017cnn}. Using $d^\prime$ allows us to better compare between datasets with different prior distributions, since mAP is confounded with the evaluation set class prior. 
We perform macro-averaging before converting to $d^\prime$ since, as probabilities, AUCs may be meaningfully averaged. Averaging after converting to $d^\prime$ gives different results.

\subsection{Hyperparameter and checkpoint selection}
We computed evaluation metrics on our held-out validation set every 20 minutes, roughly every 3000 steps. We smoothed the raw mAP and $d^\prime$ traces at each learning rate with a 7-point moving average and picked the checkpoint centered on the highest value for each metric. %
We chose the learning rate and best checkpoint with highest $d^\prime$ on the validation set. For both evaluation sets, we report all metrics using that checkpoint. %

\subsection{Evaluation of external model}
\label{AST eval}
We evaluated the AST model\footnote{We downloaded the best single model without weight-averaging from github.com/YuanGongND/ast/tree/master/pretrained\_models, %
labeled \textit{Full AudioSet, 10 tstride, 10 fstride, without Weight Averaging, Model 1 (0.450 mAP)}} published in \cite{gong21b_interspeech}
by running the PyTorch code from the GitHub repository for that paper using the current version of the AudioSet dataset. We used the mAP and AUC computed there, and computed $d^\prime$ from the macro-averaged AUC as described above.

\section{Results}

\begin{figure}[t]
\centering
\includegraphics[scale=1.0]{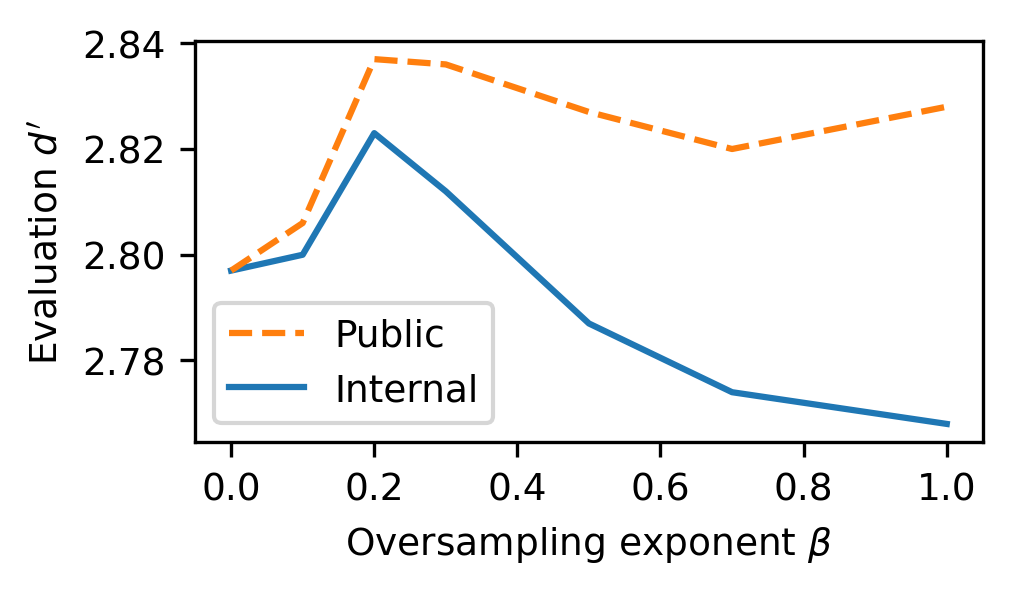}
\caption{$d^\prime$ as a function of oversampling.}
\label{fig:dprime}
\end{figure}

\subsection{Full balancing hurts performance on internal eval}
Tables \ref{tab:d_prime} and \ref{tab:MAP} show the effects of balancing on $d^\prime$ and mAP, respectively. Full balancing resulted in higher $d^\prime$ and mAP than no balancing on the public evaluation set. In contrast to this, full balancing {\em reduced} both metrics on the internal evaluation set. Partial balancing performed better than full balancing in both cases.

The public and internal evaluation sets differ in multiple ways. They clearly have different class prior distributions (Gini coefficients 0.39 and 0.61, respectively), but the added samples could substantially alter the difficulty since they were chosen from candidates for each label that were rated as ``not present''.
The similarity of the observed, prior-invariant $d^\prime$ values in Table \ref{tab:d_prime} suggests, however, that the difficulty of the two evaluations is comparable.

\subsection{Optimum balancing depends on evaluation set and metric}
We investigated the best balancing scheme by computing evaluation metrics at several values of $\beta$. Figure \ref{fig:MAP} shows the observed dropoff of internal evaluation mAP from $\beta=0$ to $\beta=1$, and the corresponding increase in public evaluation mAP. The best mAP was at $\beta=0.5$ for the public evaluation set and $\beta=0.3$ for the internal. 

Figure \ref{fig:dprime} shows a slightly different picture: $d^\prime$ was lower on the internal set with full balancing, and higher on the external set. The optimal balancing on both evaluation sets was at $\beta=0.3$

The validation set has a different class balance from the evaluation set
and might bias our results.
In practice, however, we found that the optimal validation checkpoint and learning rate were close to the oracle best values for the evaluation set.

\subsection{Balancing effects show little relation to class prior}
\begin{figure}[t]
\centering
\includegraphics[scale=1.0]{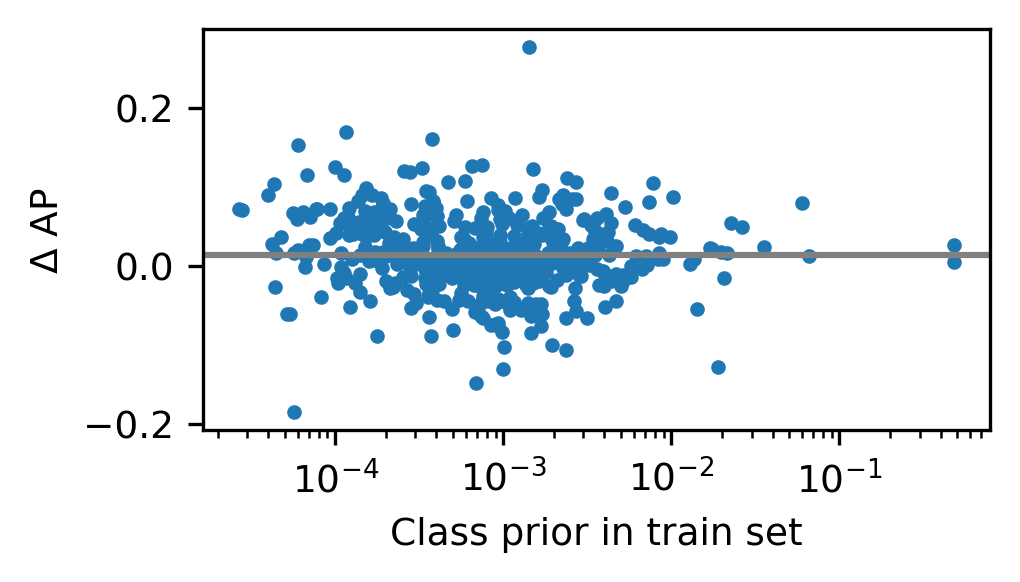}
\caption{Classwise AP improvement due to balancing vs. class prior in the training set. $\Delta$AP is computed as the difference between the AP achieved with full oversampling ($\beta=1$) and the AP achieved without oversampling ($\beta=0$). The grey line shows the mean improvement, +0.0142.}
\label{fig:deltaMAPprior}
\end{figure}

Balancing could allow the model to focus more on rare classes by presenting them more often. If this were this true, we would expect to see a greater benefit from balancing for rare classes. When we compared the per-class evaluation metrics between the models trained on the balanced ($\beta=0$) and unbalanced ($\beta=1$) training sets, the changes in per-class metrics were not correlated with the prior of the class in the training set, $p>0.1$ for the regression of ${\Delta}\textrm{AP}_k$ vs. $\log_{10}(p_k)$ (see Figure \ref{fig:deltaMAPprior}). %

\begin{table}[t]
    \centering
    \begin{tabular}{c|c|c|c|c|c}
        \multirow{3}{6mm}{}             & \multirow{3}{*}{$\beta$}  & \multicolumn{4}{c}{$d^\prime$} \\
        {}                              & {}                        & \multicolumn{2}{c}{Public} \vline & \multicolumn{2}{c}{Internal}  \\
        {}                              & {}                        & Abs.          & Rel.              & Abs.          & Rel.          \\
        \hline
        {AST}                           & {1.0}                     & 2.760         & -                 & 2.683         & -             \\
        \hline
        \multirow{3}{6mm}{This paper}   & 0.0                       & 2.797         & 0.000             & \bf{2.797}    & \bf{0.000}    \\
        {}                              & 0.5                       & \bf{2.849}    & \bf{+0.052}       & 2.787         & -0.010        \\
        {}                              & 1.0                       & 2.828         & +0.031            & 2.768         & -0.029        \\
    \end{tabular}
    \caption{
        $d^\prime$ computed on internal and public evaluation sets. $\beta$ column indicates the oversampling exponent. Relative values of $d^\prime$ are with respect to a model trained without dataset balancing and we provide them only for our models. The row labeled \emph{AST} shows the results from evaluating the best single model without weight averaging from \cite{gong21b_interspeech} as detailed in section \ref{AST eval}.}
    \label{tab:d_prime}
\end{table}

\begin{table}[t]
    \centering
    \begin{tabular}{c|c|c|c|c|c}
        \multirow{3}{6mm}{}             & \multirow{3}{*}{$\beta$}  & \multicolumn{4}{c}{mAP} \\
        {}                              & {}                        & \multicolumn{2}{c}{Public} \vline & \multicolumn{2}{c}{Internal}  \\
        {}                              & {}                        & Abs.      & Rel.                  & Abs.      & Rel.          \\
        \hline
        {AST}                           & {1.0}                     & 0.4505    & -                     & 0.3380    & -             \\
        \hline
        \multirow{3}{6mm}{This paper}   & 0.0                       & 0.4210    & 0.0000                & 0.3297    & 0.0000        \\
        {}                              & 0.5                       & 0.4394    & \bf{+0.0184}          & 0.3318    & \bf{+0.0021}  \\
        {}                              & 1.0                       & 0.4352    & +0.0142               & 0.3249    & -0.0048       \\
    \end{tabular}
    \caption{
        As Table~\ref{tab:d_prime}, but for mAP instead of $d^\prime$.
}
    \label{tab:MAP}
\end{table}

\section{Discussion}
The performance of state-of-the-art training techniques and transformer models depends heavily on both the metrics and the dataset used for evaluation. Our results show that dataset balancing does not always improve model performance, and that there is likely an interaction between the balancing scheme and the class prior distribution in the evaluation set. As shown in Table \ref{tab:d_prime}, the model from \cite{gong21b_interspeech} had lower $d^\prime$ on the internal evaluation set than the public set. This suggests that some portion of the training setup is better matched to the public evaluation set. %
The fully-balanced models from this paper shows a similar drop in $d^\prime$ from public to internal evaluation sets, -0.060 (2.828 to 2.768), compared to -0.077 for the published model (2.760 to 2.683), confirming that balancing explains the decreased performance. The model of \cite{gong21b_interspeech} performs better than our model on MAP. MAP and $d^\prime$ are sensitive to different parts of the class prior distribution, and we believe that the AST model is particularly well-tuned for MAP. 

We found qualitatively similar results when using the ImageNet ViT model \cite{dosovitskiy2020image} used by \cite{gong21b_interspeech} for pretraining instead of our audio pretraining. We observed better evaluation performance with partial balancing, and we observed worse results on the internal evaluation set with full balancing. This suggests that these phenomena are not artifacts of the audio pretraining.

Our experiments showed that balancing significantly accelerated convergence in the fine-tuning regime. The peak mAP on the validation set for the fully-balanced training occurred after 33M examples, while the peak for the unbalanced training occurred after 104M examples. Each class reached peak performance at around the same step rather than varying with prior. \cite{kong2020panns} showed that this might be possible, but did not train the models to convergence.

It is possible that this accelerated training convergence with balancing explains the 0.083 mAP difference from removing balancing shown in Table XII of \cite{gong2021psla}. That paper trained for a fixed number of epochs that we presume was optimal for full balancing. We believe that it is possible that a model trained to convergence without balancing would have seen a much smaller decrement in mAP, much closer to our result of 0.014.

We did not find meaningful correlation between per-class performance changes and the prior of the class. When performance increased or decreased it did so on average for all classes regardless of prior. Intuitively, we expected that increased representation of rare classes in training should disproportionately improve model performance for those classes, but our measurements did not back this up.

The equivocal impact of class balancing raises the question: why {\em doesn't} it work?  More training examples will, in general, improve classifier performance, yet increased representation of minority classes shows minimal benefit in Figure \ref{fig:deltaMAPprior}.  One point is that although balancing presents rare classes more often, we are not in fact increasing the diversity of those examples, but simply repeating them (with augmentation).  These large deep nets have so many parameters that perhaps they are effectively fully learning from the examples provided--even the rare ones--in the unbalanced regime, so balancing does not confer significant additional benefit.

It is possible to create other balancing schemes than the one we have described. The use, degree, and method of oversampling should be viewed as a training hyperparameter. We recommend adjusting the balancing based on performance on a held-out validation set for best generalization and to avoid overfitting to the test set.

We also observe that, since the magnitude and polarity of metric change is dependent on what should be insignificant details of the evaluation set, one should be cautious when interpreting these kinds of effects on AudioSet.

\bibliographystyle{IEEEbib}
\bibliography{citation}

\end{document}